\documentclass[12pt, oneside, a4paper]{article}

\usepackage[utf8]{inputenc}
\usepackage[sectionbib, round]{natbib}
\usepackage[textwidth = 18cm, top = 2cm, bottom = 2cm]{geometry}
\usepackage[margin=1cm]{caption}
\usepackage{amsmath, amssymb}
\usepackage{graphicx}
\usepackage{hyperref}
\usepackage{upgreek}

\RequirePackage{fancyvrb}
\RequirePackage{alltt}

\DefineVerbatimEnvironment{example}{Verbatim}{}
\renewenvironment{example*}{\begin{alltt}}{\end{alltt}}

\newcommand{\COV}{\operatorname{cov}}
\newcommand{\COR}{\operatorname{cor}}

\newcommand{\pkg}[1]{\textbf{#1}}
\newcommand{\code}[1]{\path{#1}}
\newcommand{\VEC}[1]{\boldsymbol{#1}}
\newcommand{\THETA}{\VEC{\Theta}}
\newcommand{\SIGMA}[1]{\VEC{\Sigma}_{#1}}
\newcommand{\detSIGMA}[1]{\operatorname{det} (\VEC{\Sigma}_{#1})}
\newcommand{\diagSIGMA}[1]{\operatorname{diag} (\VEC{\Sigma}_{#1})}
\newcommand{\diagCOV}[1]{\operatorname{diag} (\COV_{#1})}
\newcommand{\CRANpkg}[1]{\href{http://CRAN.R-project.org/package=#1}{\pkg{#1}}}%

\begin{document}
\bibliographystyle{abbrvnat}
\title{Multivariate Normal Mixture Modeling, Clustering and Classification with the \pkg{rebmix} Package}
\author{Marko Nagode}

\date{\today}
\maketitle

\abstract{
The \pkg{rebmix} package provides R functions for random univariate and multivariate finite mixture model generation, estimation, clustering and classification. The paper is focused on multivariate normal mixture models with unrestricted variance-covariance matrices. The objective is to show how to generate datasets for a known number of components, numbers of observations and component parameters, how to estimate the number of components, component weights and component parameters and how to predict cluster and class membership based upon a model trained by the REBMIX algorithm. The accompanying plotting, bootstrapping and other features of the package are dealt with, too. For demonstration purpose a multivariate normal dataset with unrestricted variance-covariance matrices is studied.
}

\section{Introduction}\label{sec:introduction}
Mixture models are a commonly employed tool in statistical modeling. There are quite some packages in the R statistical environment enabling number of components, component weights and component parameter estimation, clustering, classification or all \citep{Leisch_2016, Scrucca_2016}. The normal mixture models are implemented, e.g., in the following packages listed in the chronological order: \CRANpkg{mclust} \citep{Fraley_2002, Fraley_2007}, \CRANpkg{flexmix} \citep{Leisch_2004, Grun_2007, Grun_2008}, \CRANpkg{mixtools} \citep{Benaglia_2009}, \CRANpkg{HDclassif} \citep{Berge_2012}, \CRANpkg{bgmm} \citep{Przemys_2012}, \CRANpkg{EMCluster} \citep{EMCluster_20__}, \CRANpkg{Rmixmod} \citep{Lebret_2015}, \CRANpkg{mixture} \citep{Ryan_2014} and \CRANpkg{GMCM} \citep{Bilgrau_2016}. Some of them can also handle non-normal components and they all rely on the expectation-maximization (EM) algorithm or on one of its variants.

The paper aims to present the \CRANpkg{rebmix} package. REBMIX stands for the Rough and Enhanced component parameter estimation that is followed by the Bayesian classification of the remaining observations for the finite MIXture estimation. REBMIX originating in \citep{Nagode_1998} is an alternative algorithm to parameter estimation for finite mixture models. It is based on the hypothesis that the problem of finite mixture estimation can be broken into multiple problems of component parameter estimations for basic parametric family types. Here the governing equations preexist and originate in the maximum likelihood. REBMIX thus estimates component weight and component parameters for each component individually as distinguished from EM where component weights and component parameters for all components are estimated simultaneously.

The rest of the paper is organized as follows. First, the REBMIX algorithm with its input and output arguments is presented. Second, the set of equations for multivariate normal mixture models with unrestricted variance-covariance matrices is derived. Next, a multivariate normal dataset is studied to demonstrate how the \pkg{rebmix} package can be applied in the area of finite mixture estimation, clustering and classification. The last section is the conclusion.

\section{Algorithm}\label{sec:algorithm}
Let $\VEC{y}_{1}, \ldots, \VEC{y}_{n}$ be an observed $d$~dimensional dataset of size $n$ of continuous
vector observations $\VEC{y}_{j} = (y_{1j}, \ldots, y_{ij}, \ldots, y_{dj})^\top$. Each observation is assumed to follow the predictive mixture density
\begin{equation}
f(\VEC{y} | c, \VEC{w}, \THETA) = \sum_{l = 1}^{c} w_{l} f(\VEC{y} | \VEC{\uptheta}_{l})
\end{equation}
with multivariate normal component densities
\begin{equation}
f(\VEC{y} | \VEC{\uptheta}_{l}) = \frac{1}{\sqrt{(2 \pi)^{d} \detSIGMA{l}}}  \exp \left \{-\frac{1}{2} (\VEC{y} - \VEC{\mu}_{l})^\top \SIGMA{l}^{-1} (\VEC{y} - \VEC{\mu}_{l}) \right \}.
\end{equation}
The objective is to obtain the number of components $c$, component weights $w_{l}$ summing to 1 and component parameters $\VEC{\uptheta}_{l} = (\VEC{\mu}_{l}, \SIGMA{l})^\top$. The REBMIX algorithm for conditionally independent normal, lognormal, Weibull, gamma, binomial, Poisson and Dirac component densities is explained in detail in \citet{Nagode_Fajdiga_2011a, Nagode_Fajdiga_2011b, Nagode_2015}. The paper is therefore focused on the mixtures of multivariate normal component densities with unrestricted variance-covariance matrices.

\subsection{Algorithm flow}\label{subsec:algorithm_flow}
\begin{figure}[h]\centering
\includegraphics{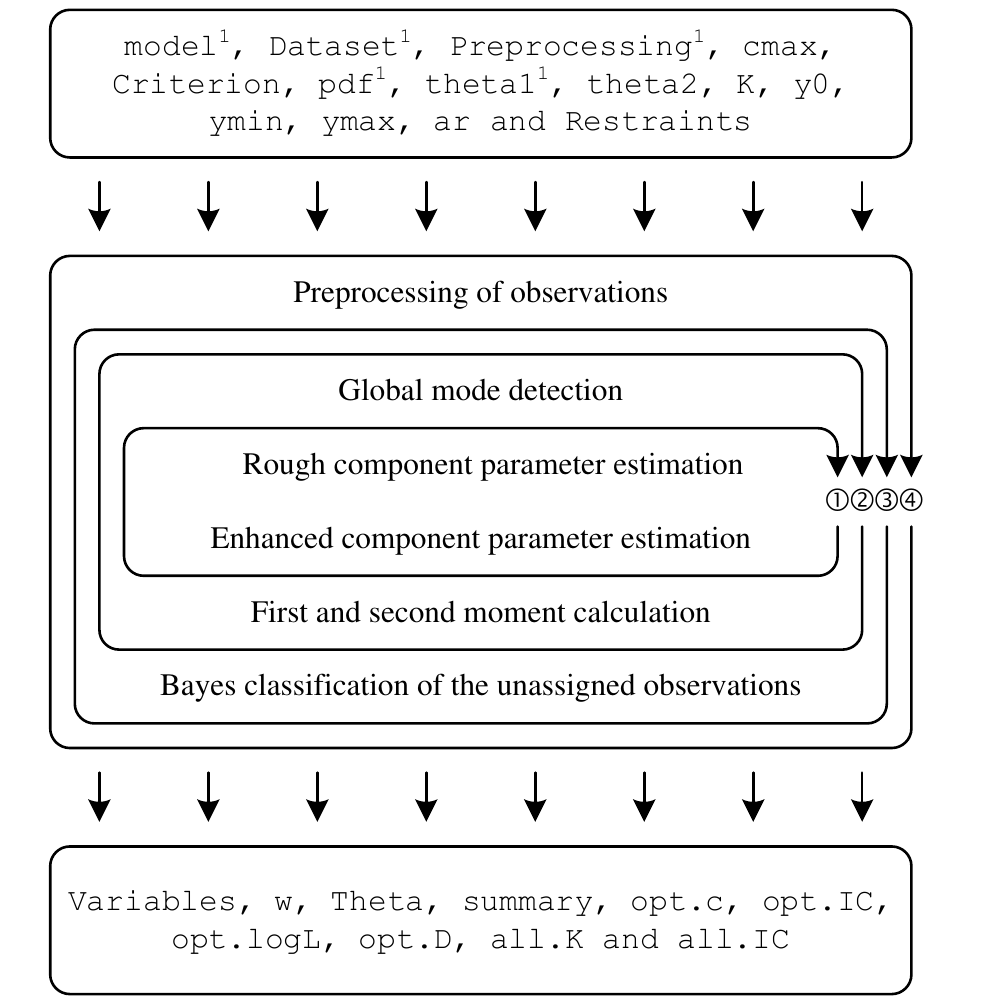}
\caption{REBMIX algorithm}\label{figure:1}
\end{figure}
\footnotetext[1]{Mandatory argument.}
The REBMIX algorithm implemented in R package \pkg{rebmix} is depicted in Fig.~\ref{figure:1}. It may be affected by maximum fourteen input arguments. Depending on the parametric families, three or four of them are mandatory while the rest are optional. For mixtures of conditionally independent component densities, argument \code{model} can be omitted. If \code{model = "REBMVNORM"}, then arguments \code{pdf} and \code{theta1} are superfluous and the output for mixtures of multivariate normal component densities with unrestricted variance-covariance matrices is returned. \code{Dataset} is a list of data frames of size $n \times d$ containing $d$~dimensional datasets. Each of the $d$ columns represents one random variable. Number of observations $n$ equals the number of rows in the datasets. \code{Preprocessing} is a character vector giving one of the preprocessing types \code{"histogram"}, \code{"Parzen window"} or \code{"k-nearest neighbour"}. Maximum number of components $c_{\mathrm{max}} \in \mathbb{N}$. The default value is \code{cmax = 15}. \code{Criterion} is a character vector giving one of the information criterion types default Akaike \code{"AIC"}, \code{"AIC3"}, \code{"AIC4"} or \code{"AICc"}, Bayesian \code{"BIC"}, consistent Akaike \code{"CAIC"}, Hannan-Quinn \code{"HQC"}, minimum description length \code{"MDL2"} or \code{"MDL5"}, approximate weight of evidence \code{"AWE"}, classification likelihood \code{"CLC"}, integrated classification likelihood \code{"ICL"} or \code{"ICL-BIC"}, partition coefficient \code{"PC"}, total of positive relative deviations \code{"D"} or sum of squares error \code{"SSE"}. Argument \code{pdf} is a character vector of length $d$ composed of one of continuous or discrete parametric family types \code{"normal"}, \code{"lognormal"}, \code{"Weibull"}, \code{"gamma"}, \code{"vonMises"}, \code{"binomial"}, \code{"Poisson"} or \code{"Dirac"}. Initial component parameters \code{theta1} is a vector of length $d$ and equals $n_{il} = \textrm{Number of categories} - 1$ for \code{"binomial"} distribution or \code{"NA"} otherwise. Argument \code{theta2} is a vector of length $d$ and is currently not used.

\code{K} is a vector or a list of vectors containing numbers of bins $v$ for the histogram and the Parzen window or numbers of nearest
neighbours $k$ for the $k$~nearest neighbour. There is no genuine rule to identify $v$ or $k$. Consequently, the algorithm identifies them from the set \code{K} of input values by minimizing the information criterion. The \citet{Sturges_1926} rule $v = 1 + \log_{2}(n)$, $\mathrm{Log}_{10}$ rule $v = 10 \log_{10}(n)$ or RootN
rule $v = 2 \sqrt{n}$ can be applied to estimate the limiting numbers of bins or the rule of thumb $k = \sqrt{n}$ to guess the intermediate number of nearest neighbours. If, e.g., \code{K = c(10, 20, 40, 60)} and minimum \code{IC} coincides with \code{40}, brackets are set to \code{20} and \code{60} and the golden section is applied to refine the minimum search. The default value is \code{"auto"}. Arguments \code{y0}, \code{ymin} and \code{ymax} are vectors of length $d$ that hold origins, minimum and maximum observations respectively. The default values are \code{numeric()}. Argument \code{ar} stands for acceleration rate $0 < a_{\mathrm{r}} \leq 1$. The default value is \code{0.1} and does not have to be altered. \code{Restraints} is a character giving one of the restraints type \code{"rigid"} or default \code{"loose"}. The rigid restraints are obsolete and applicable for well separated components only.

The algorithm is to a large extent independent of parametric families. This means that when it is extended to other parametric families, only the set of equations for rough and enhanced component parameter estimation, first and second moment calculation and Bayes classification of the unassigned observations have to be derived. The rest remains untouched.

An object of class \code{"REBMIX"} or \code{"REBMVNORM"} is returned by the \code{REBMIX} method. The object includes input and output arguments. Output \code{Variables} is a character vector of length $d$ containing one of the types of variables \code{"continuous"} or \code{"discrete"}. Argument \code{w} is a list of vectors of length $c$ that holds component weights $w_{l}$ summing to 1. \code{Theta} is a list of lists each containing $c$ parametric family types \code{pdfl}. Each character vector \code{pdfl} is composed of \code{"normal"}, \code{"lognormal"}, \code{"Weibull"}, \code{"gamma"}, \code{"vonMises"}, \code{"binomial"}, \code{"Poisson"} or \code{"Dirac"} parametric family types. Component parameters \code{theta1.l} follow the parametric family types. Each vector \code{theta1.l} is composed of $\mu_{il}$ for normal, lognormal and von Mises distributions or $\theta_{il}$ for Weibull, gamma, binomial, Poisson and Dirac distributions. Component parameters \code{theta2.l} follow \code{theta1.l}. Each vector \code{theta2.l} is composed of $\sigma_{il}$ for normal and lognormal distributions, $\beta_{il}$ for Weibull and gamma distributions, $p_{il}$ for binomial distribution and $\kappa_{il}$ for von Mises distribution. Argument \code{summary} is a data frame with additional information about dataset, preprocessing, $c_{\mathrm{max}}$, information criterion type, $a_{\mathrm{r}}$, restraints type, optimal $c$, optimal $v$ or $k$, $K$, $y_{i0}$, $y_{i\mathrm{min}}$, $y_{i\mathrm{max}}$, optimal bin widths $h_{i}$, information criterion $\mathrm{IC}$, log likelihood $\log L$ and degrees of freedom $M$.

Output arguments \code{opt.c}, \code{opt.IC}, \code{opt.logL} and \code{opt.D} are lists of vectors that contain numbers of components, information criteria, log likelihoods and totals of positive relative deviations for optimal $v$ for the histogram and the Parzen window or for optimal number of nearest neighbours $k$ for the $k$~nearest neighbour. Output \code{all.K} and \code{all.IC} are lists of vectors with all processed numbers of bins $v$ for the histogram and the Parzen window or for all processed numbers of nearest neighbours $k$ for the $k$~nearest neighbour and the corresponding information criteria.

\subsection{Preprocessing of observations}\label{subsec:preprocessing_of_observations}
The observations have to be preprocessed initially. In other words, empirical densities have to be assigned to the observations. For this purpose the histogram, Parzen window or $k$~nearest neighbour may be used. Here number of bins $v$ or nearest neighbours $k$ plays an important role. To find the number of bins or nearest neighbours resulting in the lowest value of the information criterion, the loop~4 in Fig.~\ref{figure:1} runs for all \code{K}. Package \CRANpkg{KernSmooth} may, e.g., be used to attain the optimal number of bins for $d = 1$.

\subsection{Global mode detection}\label{subsec:global_mode_detection}
Fundamental assumption of the REBMIX algorithm is that at least one component of the mixture should appear at the vicinity of the global mode, where for the Parzen window and $k$~nearest neighbour the global mode stands for the $d$~dimensional observation $\VEC{y}_{m}$ with the highest empirical density. For the histogram the global mode corresponds to the mean of a bin $\bar{\VEC{y}}_{m}$ with the highest empirical density. Index $m$ is used here to denote this particular value in the $d$~dimensional space. The loops~3 and 2 in Fig.~\ref{figure:1} are executed iteratively. The algorithm presumes only one component initially and calculates the information criterion for a mixture with that component. Onwards the number of components increases gradually and the information criterion is calculated for the corresponding mixtures. The loops~3 and 2 stop when number of components $c \geq c_{\mathrm{max}}$ or $c \geq v$ or $c \geq k$.

\subsection{Rough component parameter estimation}\label{subsec:rough_component_parameter_estimation}
Rough multivariate normal component parameter estimation is based on the component conditional densities
\begin{equation}
f(y_{i} | \VEC{y}_{\hat{i}}, \VEC{\uptheta}_{il}) = \frac{1}{\sqrt{2 \pi} \sigma_{il}} \exp \left \{ -\frac{1}{2} \frac{(y_{i} - \mu_{il})^2)}{\sigma_{il}^{2}}\right \}, \ i = 1, \ldots, d.
\end{equation}
If the equality of the predictive and empirical component conditional densities
\begin{equation}
f(\hat{y}_{im} | \hat{\VEC{y}}_{\hat{i}m}, \VEC{\uptheta}_{il}) = f_{i | \hat{i}.lm}
\end{equation}
at the global mode $\hat{\VEC{y}}_{m}$ is met, then
\begin{equation}
\mu_{il} = \hat{y}_{im} \textrm{, }  \sigma_{il} = \frac{1}{\sqrt{2 \pi} f_{i | \hat{i}.lm}}.
\end{equation}
Index $\hat{i} = 1, \ldots, i - 1, i + 1, \ldots, d$. Observation $\hat{\VEC{y}}_{m}$ equals the mean of a bin $\bar{\VEC{y}}_{m}$ for the histogram and $\hat{\VEC{y}}_{m} = \VEC{y}_{m}$ for the Parzen window and $k$~nearest neighbour \citep{Nagode_2015}. For the histogram, the variance-covariance matrix is given by
\begin{equation}
\COV_{l} = \frac{1}{n_{l}} \sum_{j = 1}^{v} k_{lj} (\hat{\VEC{y}}_{j} - \VEC{\mu}_{l}) (\hat{\VEC{y}}_{j} - \VEC{\mu}_{l})^\top,
\end{equation}
where $k_{lj}$ denote frequencies and $\VEC{\mu}_{l} = (\mu_{1l}, \ldots, \mu_{dl})^\top$. Number of bins $v$ is replaced by $n$ for the Parzen window or $k$~nearest neighbour. Once the variance-covariance matrix is known, the correlation matrix
\begin{equation}
\COR_{l} = \diagCOV{l}^{-\frac{1}{2}} \COV_{l}\diagCOV{l}^{-\frac{1}{2}} = \{\rho_{i\tilde{i}l}\}
\end{equation}
and its inverse
\begin{equation}
\COR_{l}^{-1} = \{g_{i\tilde{i}l}\}
\end{equation}
are determined. It can be proved that
\begin{equation}
\sigma_{iil} = g_{iil} \sigma_{il}^2, \ i = 1, \ldots, d,
\end{equation}
which yields
\begin{equation}
\SIGMA{l} = \diagSIGMA{l}^{\frac{1}{2}} \COR_{l} \diagSIGMA{l}^{\frac{1}{2}} = \{\sigma_{i\tilde{i}l}\}.
\end{equation}
The idea is to prevent the component from flowing away from the global mode as at least one component is supposed to be in its vicinity. This yields
\begin{equation}
f(\VEC{y} = \hat{\VEC{y}}_{m} | \VEC{\uptheta}_{l}) = \frac{1}{\sqrt{(2 \pi \varepsilon)^{d} \detSIGMA{l}}} = f_{lm},
\end{equation}
wherefrom
\begin{equation}
\varepsilon = \mathrm{max} \left \{1, \left (f_{lm} \sqrt{(2 \pi)^{d} \detSIGMA{l}} \right )^{-\frac{2}{d}} \right \}.
\end{equation}
The lower limit of $\varepsilon$ is set to $1$ in order to prevent $f(\VEC{y} = \hat{\VEC{y}}_{m} | \VEC{\uptheta}_{l})$ to be less than the empirical density at the global mode $f_{lm}$. This finally yields
\begin{equation}
\SIGMA{l} = \varepsilon \SIGMA{l} = \{\varepsilon \sigma_{i\tilde{i}l}\}.
\end{equation}
The loose restraints are treated exactly the same way as in \citet{Nagode_2015}. The loop~1 in Fig.~\ref{figure:1} splits the dataset into two clusters, the one corresponding to the currently observed component and the residue. The former is used to estimate the component weight and rough component parameters. The latter is split into two clusters repeatedly until $c \geq c_{\mathrm{max}}$ or $c \geq v$ or $c \geq k$. When the total of positive relative deviations attains its minimum, the enhanced component parameters are estimated and the loop~1 stops.

\subsection{Enhanced component parameter estimation}\label{subsec:enhanced_component_parameter_estimation}
Maximum likelihood is employed to obtain enhanced component parameters. For the histogram, enhanced multivariate normal component
parameters are given by
\begin{equation}
\VEC{\mu}_{l} = \frac{1}{n_{l}} \sum_{j = 1}^{v} k_{lj} \hat{\VEC{y}}_{j} \textrm{ and } \SIGMA{l} = \frac{1}{n_{l}} \sum_{j = 1}^{v} k_{lj} (\hat{\VEC{y}}_{j} - \VEC{\mu}_{l}) (\hat{\VEC{y}}_{j} - \VEC{\mu}_{l})^\top.
\end{equation}
Index $v$ is replaced by $n$ for the Parzen window or $k$~nearest neighbour.
\subsection{First and second moment calculation}\label{subsec:first_and_second_moment_calculation}
The first and second moment
\begin{equation}\label{eq:1}
\VEC{m}_{l} = \VEC{\mu}_{l} \textrm{ and } \VEC{V}_{l} = \SIGMA{l} + \VEC{\mu}_{l} \VEC{\mu}_{l}^\top = \{V_{i\tilde{i}l}\}
\end{equation}
of the multivariate normal distribution are required for classification of the unassigned observations.
\subsection{Bayes classification of the unassigned observations}\label{subsec:bayes_classification_of_the_unassigned_observations}
When the weight of the unassigned observations $w_{l} \leq D_{\mathrm{min}} (l - 1)$, then the assignment of new components stops. The constant $0 < D_{\mathrm{min}} \leq 1$ is optimized by one of the information criteria. Unassigned observations $k_{lj}$ are then assumed to belong to the existing classes. The classification of unassigned observations is accomplished by the Bayes decision rule \cite{Duda_and_Hart_1973}
\begin{gather}
l =  \underset{l}{\operatorname{arg} \operatorname{max}} w_{l} f(\VEC{y}_{j} | \VEC{\uptheta}_{l}) \nonumber \\ w_{l} = w_{l} + \frac{k_{lj}}{n} \textrm{, } m_{il} = m_{il} + \frac{k_{lj} (y_{ij} - m_{il})}{n w_{l}} \textrm{ and } V_{i\tilde{i}l} = V_{i\tilde{i}l} + \frac{k_{lj} (y_{ij} y_{\tilde{i}j} - V_{i\tilde{i}l})}{n w_{l}},
\end{gather}
where $k_{lj}$ is added to the $l$th class and the component weight and both moments are recalculated \cite{Bishop_1995}. Once all $v$ bin means or all $n$ observations are processed, the predictive mixture parameters are gained by inverting Equation~(\ref{eq:1}).

The overall optimal number of components, component weights and component parameters are those resulting in the lowest value of the information criterion.

\section{Example}\label{sec:examples}
For demonstration purpose a multivariate normal dataset with unrestricted variance-covariance matrices is studied. By varying dataset dimension $d$, dataset size $n$, number of components $c$, the seed, range of means $\VEC{\mu}_{l}$, range of eigenvalues $\VEC{\Lambda}_{l}$ of variance-covariance matrix $\SIGMA{l} = \VEC{P}_{l} \VEC{\Lambda}_{l} \VEC{P}_{l}^\top$ and eigenvectors stored in $\VEC{P}_{l}$, a great variety of very different datasets may be generated and the package can be tested.

\subsection{Random dataset generation}\label{subsec:random_dataset_generation}
\begin{figure}[h]\centering
\includegraphics{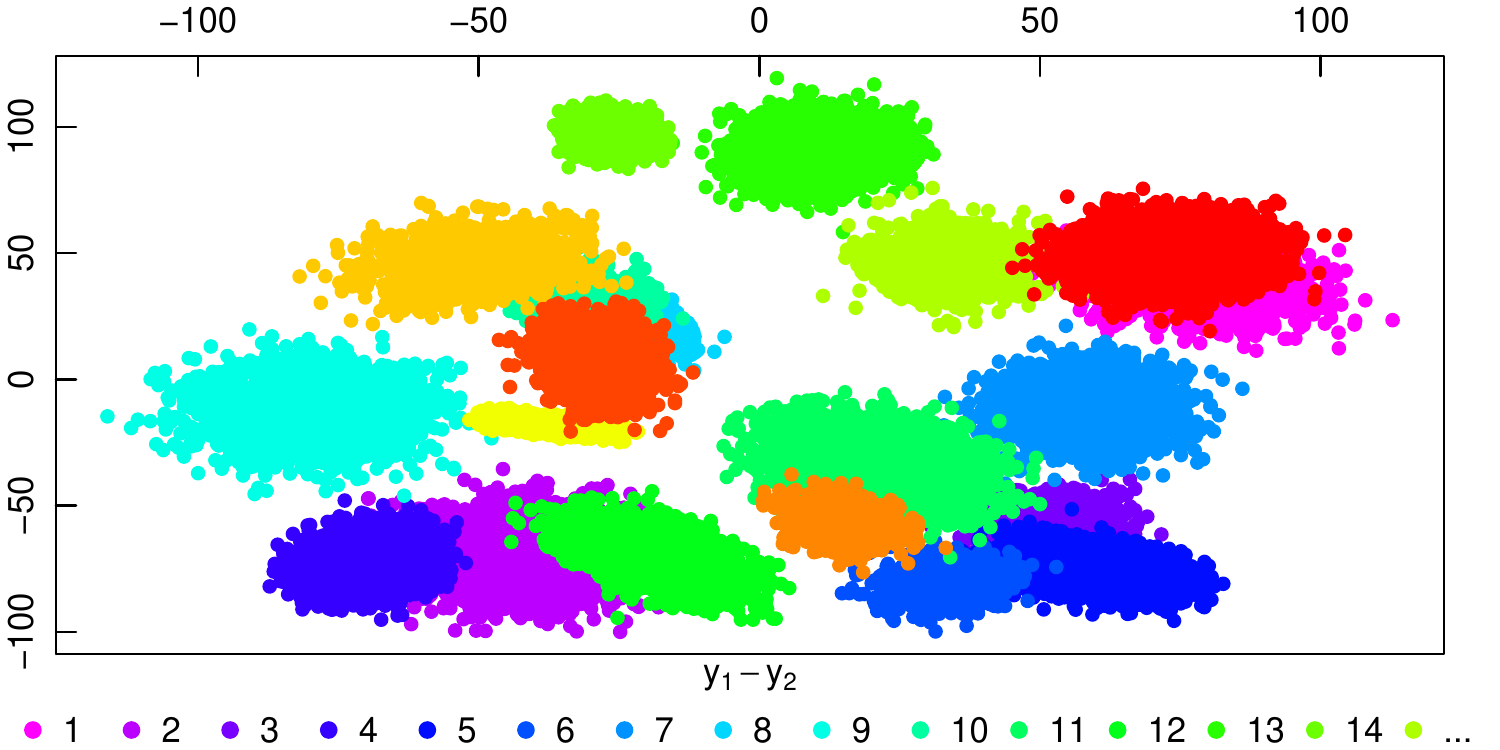}
\caption{Simulated overlapped multivariate normal dataset with known cluster membership}\label{figure:2}
\end{figure}

\begin{figure}[h]\centering
\includegraphics{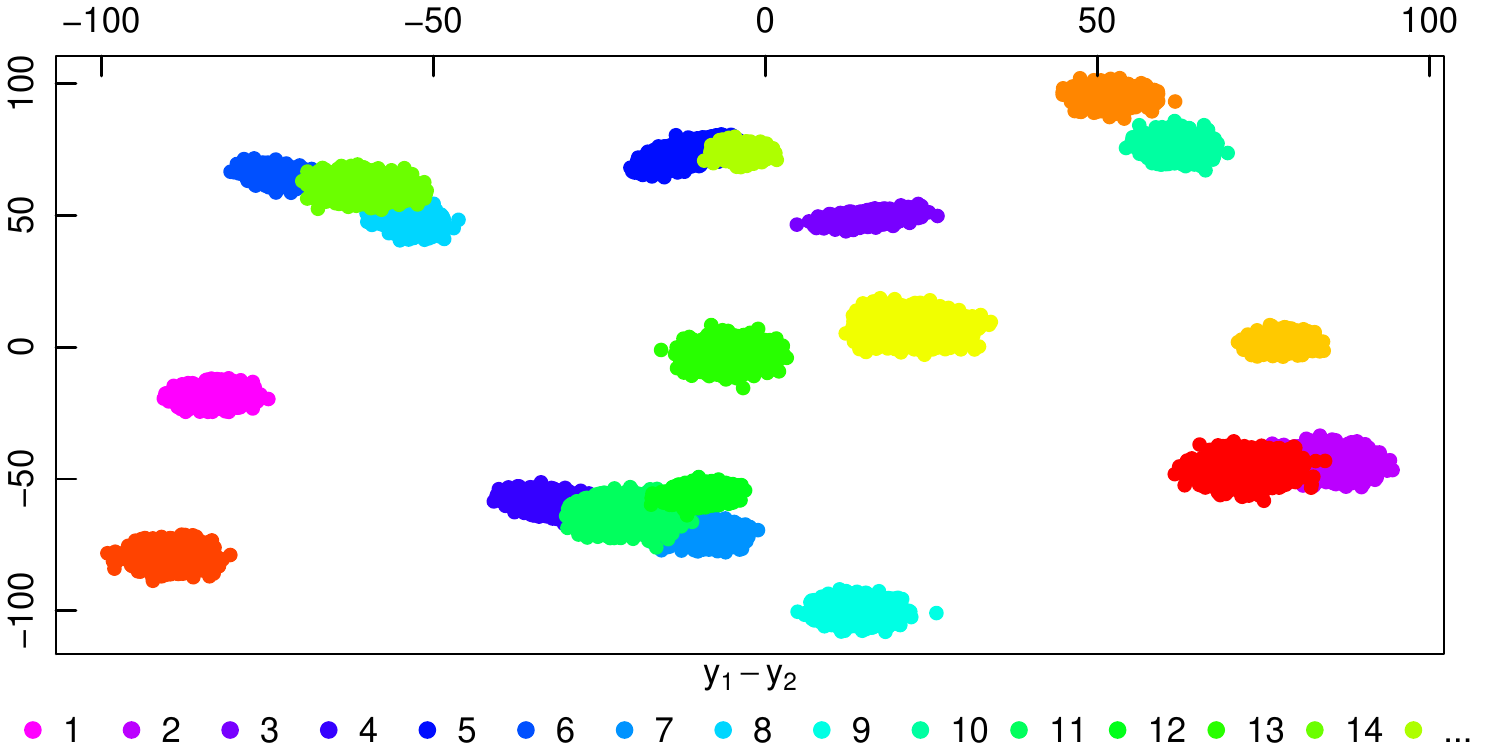}
\caption{Simulated less overlapped multivariate normal dataset with known cluster membership}\label{figure:3}
\end{figure}

This section demonstrates how multivariate normal datasets with unrestricted variance-covariance matrices are generated. Usually the \code{d} dimensional datasets of size \code{n} are generated for some chosen component weights $w_{l}$ and component parameters $\VEC{\mu}_{l}$ and $\SIGMA{l}$, where $l$ ranges from $1$ to \code{c}. Here the diversity of datasets is achieved by varying
\begin{example}
d <- 2; n <- 50000; c <- 20; set.seed(123)
\end{example}
In order to increase the diversity of datasets, the chosen component weights are replaced by randomly generated ones
\begin{example}
w <- runif(c, 0.1, 0.9); w <- w / sum(w)
\end{example}
Other parameters affecting dataset generation are ranges of means \code{mu} and eigenvalues \code{lambda}
\begin{example}
mu <- c(-100, 100); lambda <- c(1, 100)
\end{example}
Component means \code{Mu} and diagonal matrices of eigenvalues \code{Lambda} are uniformly distributed. The limits of the distributions are in \code{mu} and \code{lambda}. Component variance-covariance matrices \code{Sigma} are generated by the singular-value decomposition of uniformly distributed rectangular matrices. Here the limits of the distribution are $-1$ and $1$.
\begin{example}
Mu <- list(); Sigma <- list()
for (l in 1:c) {
  Mu[[l]] <- runif(d, mu[1], mu[2])
  Lambda <- diag(runif(d, lambda[1], lambda[2]), nrow = d, ncol = d)
  P <- svd(matrix(runif(d * d, -1, 1), nc = d))$u
  Sigma[[l]] <- P %*% Lambda %*% t(P)
}
\end{example}
Numbers of observations in classes \code{n}, \code{Mu} and \code{Sigma} are strored in a format required by the \pkg{rebmix} package
\begin{example}
n <- round(w * n); Theta <- list()
for (l in 1:c) {
  Theta[[paste0("pdf", l)]] <- rep("normal", d)
  Theta[[paste0("theta1.", l)]] <- Mu[[l]]
  Theta[[paste0("theta2.", l)]] <- as.vector(Sigma[[l]])
}
\end{example}
Now, all arguments entering the \code{RNGMIX} method are ready for finite mixture generation
\begin{example}
mvnorm <- RNGMIX(model = "RNGMVNORM", Dataset.name = "mvnorm_1",
  rseed = -1, n = n, Theta = Theta)
\end{example}
The method returns an object of class \code{"RNGMVNORM"} containing \code{Dataset.name}, \code{rseed}, \code{n}, \code{Theta}, \code{Dataset}, \code{Zt}, \code{w}, \code{Variables}, \code{ymin} and \code{ymax} slots. Refer to \code{help("RNGMIX-class")} for details. Only three slots are shown if \code{mvnorm} is called
\begin{example}
> mvnorm
An object of class "RNGMVNORM"
Slot "w":
 [1] 0.0305 0.0676 0.0395 0.0746 0.0788 0.0126 0.0483 0.0753 0.0500 0.0430
[11] 0.0800 0.0428 0.0594 0.0516 0.0169 0.0758 0.0275 0.0124 0.0335 0.0799
Slot "ymin":
[1] -116 -100
Slot "ymax":
[1] 113 119
\end{example}
To plot the dataset, call
\begin{example}
plot(mvnorm)
\end{example}
In Fig.~\ref{figure:2} the clustered multivariate normal dataset with considerably overlapped components is given. True cluster membership \code{Zt} is known. Each cluster is depicted in a different colour.
If the code of this section is called anew by only changing
\begin{example}
set.seed(124); lambda <- c(1, 10)
\end{example}
a very different and less overlapped dataset in Fig.~\ref{figure:3} is obtained.

\subsection{Finite mixture estimation}\label{subsec:finite_mixture_estimation}
This section shows how normal finite mixture with unrestricted variance-covariance matrices is estimated given that the dataset from the previous section is known. By calling the \code{REBMIX} method, finite mixture for the data frame \code{mvnorm\_1} stored in the list \code{mvnorm@Dataset} is estimated. From experience it is reasonable to set \code{Preprocessing} to \code{"histogram"} for $n > 1000$. Otherwise \code{"Parzen window"} or \code{"k-nearest neighbour"} may be preferable. The algorithm always searches the optimal number of components between $1$ and $c_{\mathrm{max}}$. Maximum number of components $c_{\mathrm{max}}$ is, as a rule, chosen to be larger than the maximum expected number of components. For the complete list of arguments refer to \code{help("REBMIX-methods")}.
\begin{example}
mvnormest <- REBMIX(model = "REBMVNORM", Dataset = mvnorm@Dataset,
  Criterion = "BIC", Preprocessing = "histogram", cmax = ceiling(1.2 * c))
\end{example}
Object \code{mvnormest} of class \code{"REBMVNORM"} contains \code{Dataset}, \code{Preprocessing}, \code{cmax}, \code{Criterion}, \code{Variables}, \code{pdf}, \code{theta1}, \code{theta2}, \code{K}, \code{y0}, \code{ymin}, \code{ymax}, \code{ar}, \code{Restraints}, \code{w}, \code{Theta}, \code{summary}, \code{pos}, \code{opt.c}, \code{opt.IC}, \code{opt.logL}, \code{opt.D}, \code{all.K} and \code{all.IC} slots. Refer to \code{help("REBMIX-class")} for details. To produce object summaries, the \code{summary} method is called.
\begin{example}
> summary(mvnormest)
   Dataset Preprocessing Criterion  c v/k     IC    logL   M
1 mvnorm_1     histogram       BIC 24  46 931191 -464822 143
Maximum logL = -464822 at pos = 1.
\end{example}
In the particular case only one dataset is stored in \code{mvnorm@Dataset}. However, it is possible to store more than one dataset to \code{mvnorm@Dataset}. All datasets are then processed with the same arguments when the \code{REBMIX} method is called. Refer to the example section of \code{help("RNGMIX-methods")} for details. To print the coefficients, the \code{coef} method is called, where \code{pos} stands for the desired row number in \code{mvnormest@summary} to be printed or plotted.
\begin{example*}
> coef(mvnormest, pos = 1)
   comp1  comp2 comp3  comp4  comp5  comp6
w 0.0328 0.0335 0.071 0.0339 0.0712 0.0124
 \vdots
              1      2
theta1.1  -38.6 -17.28
theta1.2  -38.4 -17.43
theta1.3  -24.5  21.91
theta1.4  -24.9  97.55
theta1.5  -68.3 -72.67
theta1.6  -32.6 -20.44
 \vdots
            1-1      1-2      2-1    2-2
theta2.1   27.1  -3.0596  -3.0596   5.38
theta2.2   11.1  -0.9241  -0.9241   3.79
theta2.3   24.0 -15.9239 -15.9239  82.41
theta2.4   11.7  -2.8936  -2.8936  15.97
theta2.5   42.6  13.9595  13.9595  48.01
theta2.6   10.6   3.8294   3.8294  14.21
 \vdots
\end{example*}
The \code{plot} method is called to plot the \code{mvnormest} object in Fig~\ref{figure:3}. Refer to \code{help("plot-methods")} for details.
\begin{example}
plot(mvnormest, pos = 1, nrow = 3, ncol = 1, what = c("density", "marginal"))
\end{example}
\begin{figure}[h]\centering
\includegraphics{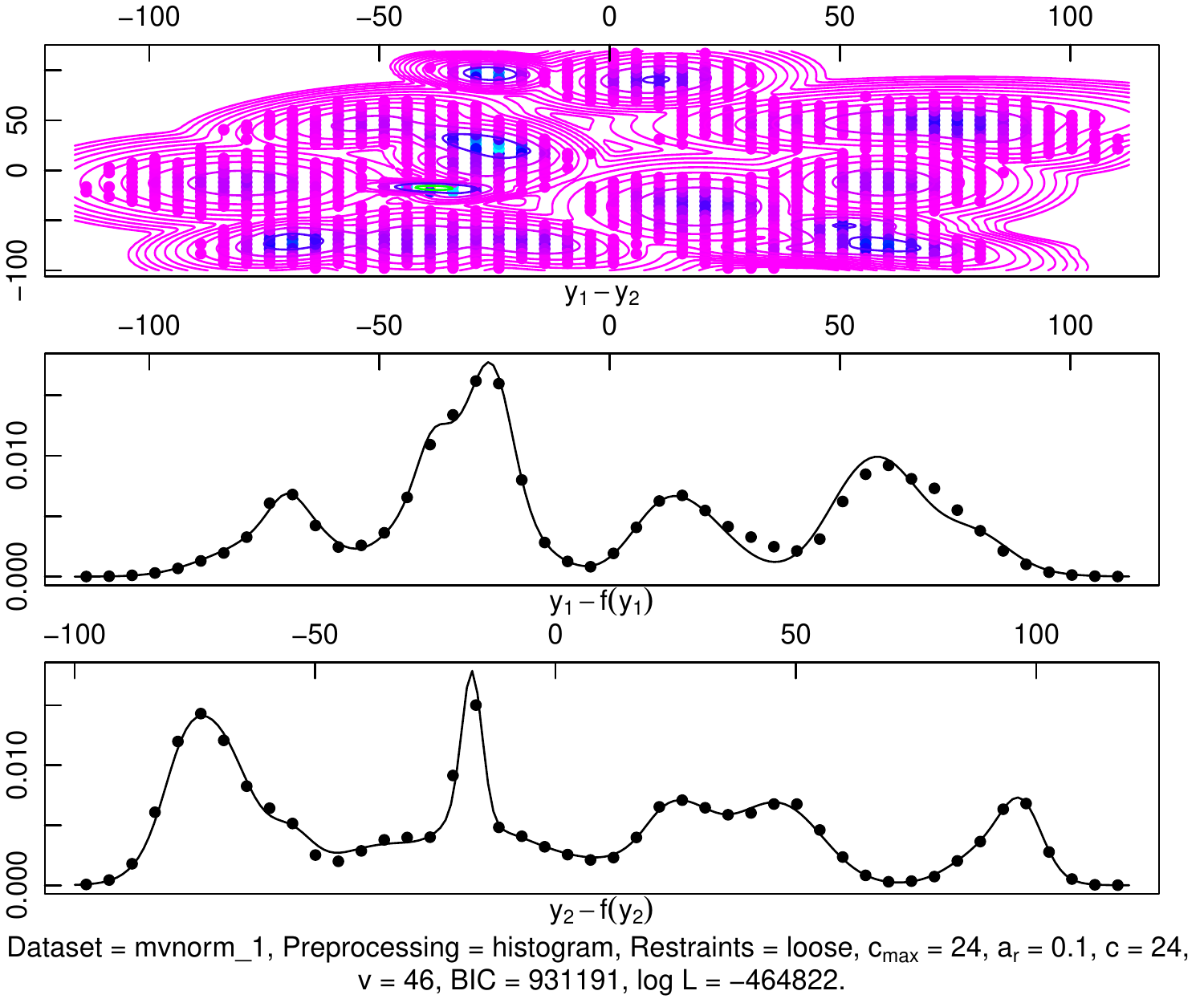}
\caption{Empirical densities (coloured circles), predictive multivariate normal mixture density (coloured lines), empirical densities (circles), predictive univariate marginal normal mixture densities (solid lines)}\label{figure:4}
\end{figure}

\subsection{Bootstrapping}\label{subsec:bootstrapping}
Bootstrapping is dealt with in this section. It is the practice of estimating standard errors and coefficients of variation by measuring those properties when sampling from approximating distribution stored in the \code{mvnormest} object. If \code{Bootstrap = "parametric"}, the bootstrap datasets are generated by executing the \code{RNGMIX} method within the \code{boot} method based on the component weights and component parameters of the \code{mvnormest} object. The number of bootstrap datasets is controlled by the \code{B} argument. If \code{Bootstrap = "nonparametric"}, the bootstrap datasets are generated directly from the \code{mvnorm\_1} data frame by applying the \code{sample.int} built-in R method. The \code{boot} method
\begin{example}
mvnormboot <- boot(mvnormest, pos = 1, Bootstrap = "parametric", B = 10)
\end{example}
returns object \code{mvnormboot} of class \code{"REBMVNORM.boot"}. It contains \code{x}, \code{rseed}, \code{pos}, \code{Bootstrap}, \code{B}, \code{n}, \code{replace}, \code{prob}, \code{c}, \code{c.se}, \code{c.cv}, \code{c.mode}, \code{c.prob}, \code{w}, \code{w.se}, \code{w.cv}, \code{Theta}, \code{Theta.se} and \code{Theta.cv} slots. Refer to \code{help("REBMIX.boot-class")} for details. Only the most important five slots are shown if \code{mvnormboot} is called.
\begin{example}
> mvnormboot
An object of class "REBMVNORM.boot"
Slot "c":
 [1] 24 23 21 23 23 23 24 22 24 23
Slot "c.se":
[1] 0.943
Slot "c.cv":
[1] 0.041
Slot "c.mode":
[1] 23
Slot "c.prob":
[1] 0.5
\end{example}
The optimal number of components in slot \code{c} is returned for each bootstrap dataset. When standard errors and coefficients of variation are calculated, only those datasets are taken into account for which \code{c} equals \code{c.mode}. In the particular case only five datasets out of ten are considered. The \code{summary} method returns coefficients of variation of component weights and component parameters. The probability of identifying exactly \code{c.mode} components is returned, too.
\begin{example*}
> summary(mvnormboot)
     comp1 comp2 comp3 comp4 comp5 comp6
w.cv  0.46 0.692 0.673 0.679 0.235 0.439
 \vdots
                    1       2
theta1.1.cv   -0.0306  -0.127
theta1.2.cv   -0.1466  -2.040
theta1.3.cv   -0.2214   2.056
theta1.4.cv   -0.2033   2.174
theta1.5.cv   -0.5526   3.060
theta1.6.cv    4.5885  -0.419
 \vdots
               1-1     1-2     2-1   2-2
theta2.1.cv  0.593   0.949   0.949 0.895
theta2.2.cv  0.534  -1.822  -1.822 1.715
theta2.3.cv  0.538  -1.194  -1.194 1.100
theta2.4.cv  0.461  -1.223  -1.223 1.026
theta2.5.cv  0.385   2.452   2.452 0.605
theta2.6.cv  0.495  -1.551  -1.551 0.567
 \vdots
Mode probability = 0.5 at c = 23 components.
\end{example*}

\subsection{Clustering}\label{subsec:clustering}
Clustering is used here to group observations in such a way that those with similar features come together and the ones with dissimilar features go apart. It belongs to unsupervised learning. Clustering requires the \code{mvnormest} object and \code{pos}. It is supposed that components exactly match the clusters and that true cluster membership \code{Zt} is unknown. In the particular multivariate normal dataset \code{Zt} is known and therefore may enter the \code{RCLRMIX} method.
\begin{example}
mvnormclu <- RCLRMIX(model = "RCLRMVNORM", mvnormest, pos = 1, Zt = mvnorm@Zt)
\end{example}
Object \code{mvnormclu} of class \code{"RCLRMIX"} is returned. It contains \code{x}, \code{pos}, \code{Zt}, \code{Zp}, \code{c}, \code{prob}, \code{from}, \code{to}, \code{EN} and \code{ED} slots. Refer to \code{help("RCLRMIX-class")} for details. Combining mixture components for clustering follows \citet{Baudry_2010}. If \code{Zt} is known and enters the \code{RCLRMIX} method, then the probabilities of correct clustering \code{mvnormclu@prob} are calculated, which is convenient to test the package. The optimal number of clusters is given by
\begin{example}
copt <- which.max(mvnormclu@prob))
> copt
[1] 16
\end{example}
and does not necessarily coincide with the true number of clusters. To plot the \code{mvnormclu} object in Fig~\ref{figure:4} for \code{copt}, the \code{plot} method is called.
\begin{example}
plot(mvnormclu, s = copt)
\end{example}
\begin{figure}[h]\centering
\includegraphics{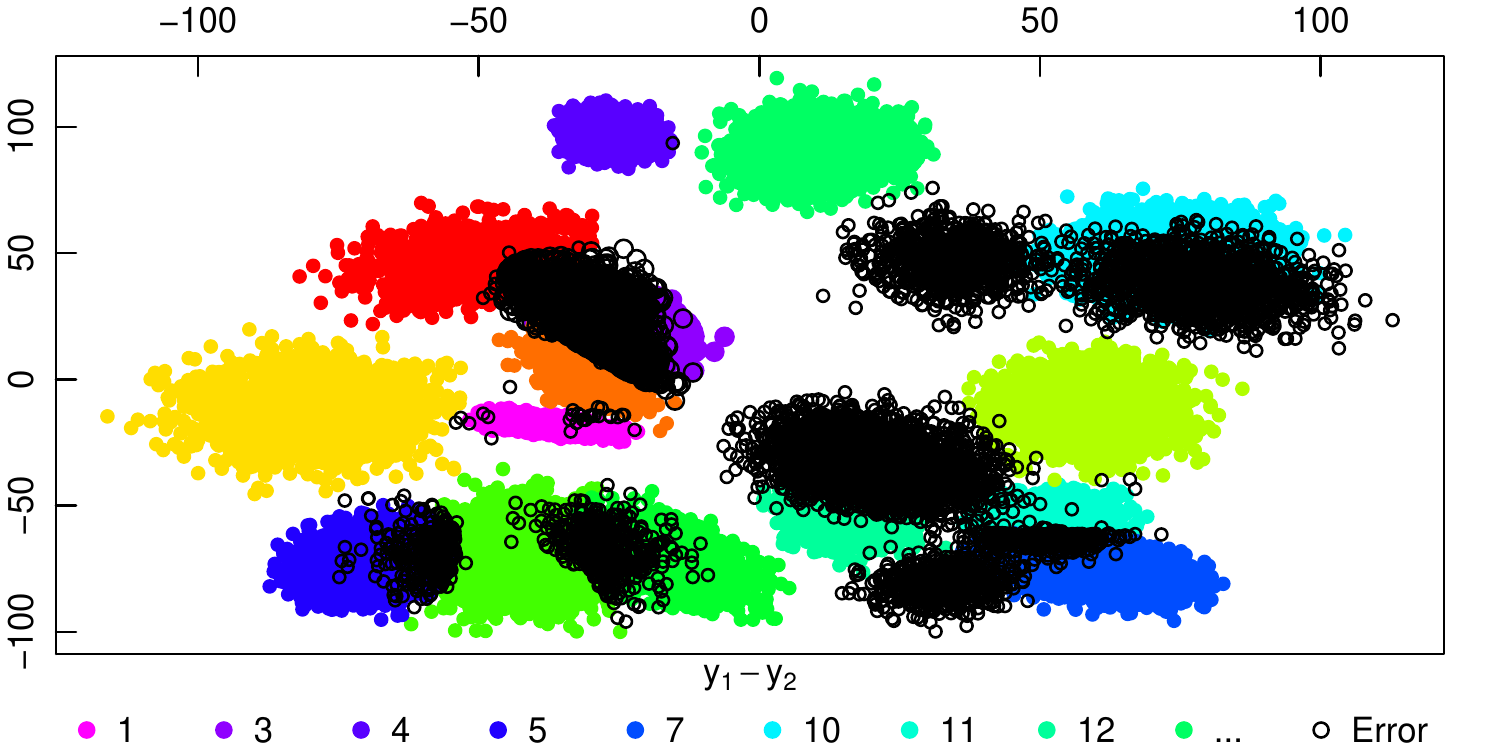}
\caption{Simulated multivariate normal dataset with predictive and error cluster membership}\label{figure:5}
\end{figure}
The probability of correct clustering for \code{copt} is then
\begin{example}
> mvnormclu@prob[copt]
[1] 0.779
\end{example}
The \code{summary} method returns entropy \code{EN} and entropy decrease \code{ED} \citep{Baudry_2010} for all possible numbers of clusters. The information on how the clusters are merged is also printed, e.g., if number of clusters equals $23$, then cluster $12$ is merged with cluster $10$.
\begin{example*}
> summary(mvnormclu)
Number of clusters  1           2          3          4          5
From cluster        4           13         5          10         7
To cluster          1           1          1          5          5
Entropy             -1.86e-13   2.39e+00   7.00e+00   1.28e+01   2.40e+01
Entropy decrease    2.39        4.61       5.79       11.21      12.15
 \vdots
Number of clusters  21         22
From cluster        19         8
To cluster          10         3
Entropy             1.21e+04   1.46e+04
Entropy decrease    2533.21    3150.78
\end{example*}

\subsection{Classification}\label{subsec:classification}
\begin{figure}[h]\centering
\includegraphics{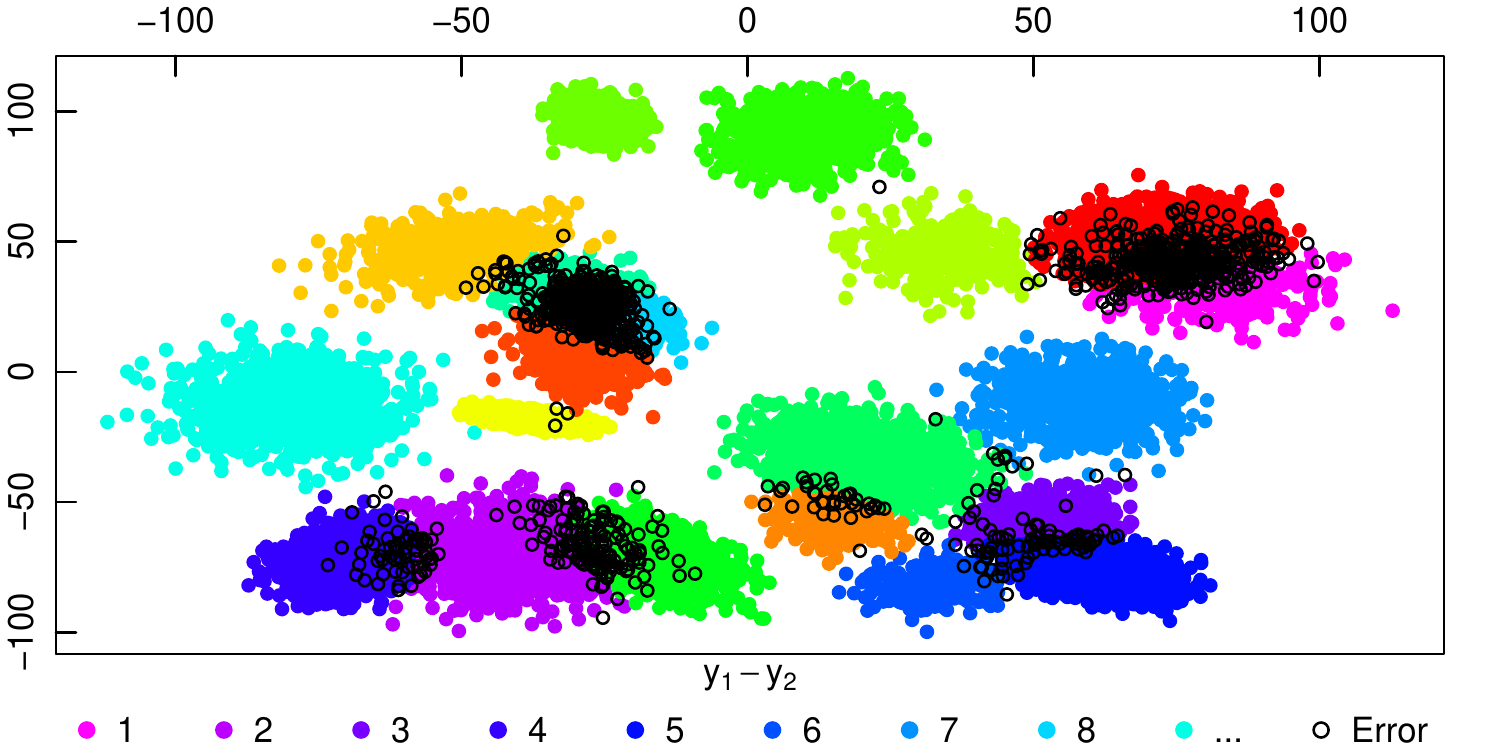}
\caption{Simulated multivariate normal dataset with predictive and error class membership}\label{figure:6}
\end{figure}

Classification is a process of categorization where observations are recognized, differentiated and understood on the basis of a train subset. It belongs to supervised learning, where for the train subset the true class membership \code{Zt} is available. The simulated multivariate normal dataset \code{mvnorm\_1} is first combined by columns with \code{Zt} from section~Random dataset generation.
\begin{example}
Dataset <- cbind(mvnorm@Zt, mvnorm@Dataset$mvnorm_1)
\end{example}
Next, \code{Dataset} is split into train $60 \%$ and test $40 \%$ datasets.
\begin{example}
Mvnorm <- split(p = 0.6, Dataset = Dataset, class = 1)
\end{example}
The \code{split} method returns the object \code{Mvnorm} of class \code{"RCLS.chunk"}. Argument \code{class} stands for the column number in \code{Dataset} containing the class membership information. Refer to \code{help("split-methods")} for details. By calling the \code{REBMIX} method, finite mixtures for the train subsets in the list \code{Mvnorm@train} are estimated. Maximum number of components \code{cmax} is set here to $5$.
\begin{example}
mvnormest <- REBMIX(model = "REBMVNORM", Dataset = Mvnorm@train,
  Preprocessing = "histogram", cmax = 5, Criterion = "BIC")
\end{example}
Classification requires the list of objects \code{mvnormest} of length $o$. In the particular example number of chunks $o$ equals $1$. \code{Dataset} is a data frame containing test dataset \code{Mvnorm@test}. Factor of true class membership \code{Zt} for the test dataset enters the \code{RCLSMIX} method, too.
\begin{example}
mvnormcla <- RCLSMIX(model = "RCLSMVNORM", x = list(mvnormest),
  Dataset = Mvnorm@test, Zt = Mvnorm@Zt)
\end{example}
Object \code{mvnormcla} of class \code{"RCLSMIX"} is returned. It contains \code{x}, \code{o}, \code{Dataset}, \code{s}, \code{ntrain}, \code{P}, \code{ntest}, \code{Zt}, \code{Zp}, \code{CM}, \code{Accuracy}, \code{Error}, \code{Precision}, \code{Sensitivity}, \code{Specificity} and \code{Chunks} slots. The \code{summary} method returns table containing confusion matrix \code{CM} for multiclass classifier and proportion of all test observations that are classified wrongly \code{Error}.
\begin{example*}
> summary(mvnormcla)
    Test Predictive Frequency
1      1          1       322
2      2          1         0
3      3          1         0
4      4          1         0
5      5          1         0
6      6          1         0
7      7          1         0
8      8          1         0
9      9          1         0
10    10          1         0
 \vdots
391   11         20         0
392   12         20         0
393   13         20         0
394   14         20         0
395   15         20         6
396   16         20         0
397   17         20         0
398   18         20         0
399   19         20         0
400   20         20      1484
Error = 0.0662.
\end{example*}
Refer to \code{help("RCLSMIX-class")} for details. To plot the \code{mvnormcla} object in Fig~\ref{figure:6}, the \code{plot} method is called.
\begin{example}
plot(mvnormcla)
\end{example}

\section{Summary}\label{sec:summary}
The \code{rebmix} package is aimed to generate, estimate, cluster and classify finite mixture models. Variables can be continuous, discrete or mixed, independent or dependent and may follow normal, lognormal, Weibull, gamma, von Mises, binomial, Poisson or Dirac parametric families. The package has been compared recently with the well established \pkg{flexmix} package by \citet{Franko_2015}. It has turned out that the \pkg{rebmix} package is especially favourable for large datasets in combination with the \code{"histogram"} preprocessing due to its numerical stability and computational speed. The preprocessing of observations is beneficial for large datasets. However, for small datasets it results in slightly poorer estimates as compared to the EM based packages. REBMIX can also be used to assess an initial set of unknown parameters and number of components, e.g., for the EM based packages.

In this paper the features for multivariate normal mixture models with unrestricted variance-covariance matrices are presented. For non-normal mixtures readers may address \citet{Nagode_Fajdiga_2011a, Nagode_Fajdiga_2011b, Nagode_2015} where they can also find thorough theoretical backgrounds of the algorithm. New features for finite mixture modeling, bootstrapping, clustering and classification are presented. The example enables the study of a great variety of very different datasets by modifying \code{d}, \code{n}, \code{c}, the seed, \code{mu} and \code{lambda}. The number specifying the fraction of observations for training \code{p} may be changed as well. All this enables further evaluations of the package.

\section{Acknowledgment}\label{sec:acknowledgment}
This paper is a part of research work within program Nr. P2-0182 entitled Development evaluation financed by the Slovenian Ministry of Education, Science and Sport.

\bibliography{nagode_article_2018}

\vspace{\baselineskip}\noindent\emph{Marko Nagode\\
University of Ljubljana\\
Faculty of Mechanical Engineering\\
A\v{s}ker\v{c}eva 6\\
1000 Ljubljana\\
Slovenia}\\
\href{mailto:Marko.Nagode@fs.uni-lj.si}{Marko.Nagode@fs.uni-lj.si}.
\end{document}